\documentclass[10pt,twocolumn,letterpaper]{article}

\usepackage{cvpr}              % To produce the CAMERA-READY version
\usepackage{graphicx}
\usepackage{amsmath}
\usepackage{amssymb}
\usepackage{booktabs}

\usepackage{multirow}
\usepackage{graphicx}

\usepackage[pagebackref,breaklinks,colorlinks]{hyperref}

\usepackage[capitalize]{cleveref}
\crefname{section}{Sec.}{Secs.}
\Crefname{section}{Section}{Sections}
\Crefname{table}{Table}{Tables}
\crefname{table}{Tab.}{Tabs.}

\usepackage{bbding}

\def\eg{{\itshape e.g.}}
\def\ie{{\itshape i.e.}}

\def\wrt{{\itshape w.r.t.}}
\def\Ours{{FMWISS}}

\def\cvprPaperID{9511} % *** Enter the CVPR Paper ID here
\def\confName{CVPR}
\def\confYear{2023}

\def\AddAppendix{AddAppendix}

\begin{document}

%%%%%%%%% TITLE - PLEASE UPDATE
\title{Foundation Model Drives Weakly Incremental Learning for Semantic Segmentation}

% \author{First Author\\
% Institution1\\
% Institution1 address\\
% {\tt\small firstauthor@i1.org}
% % For a paper whose authors are all at the same institution,
% % omit the following lines up until the closing ``}''.
% % Additional authors and addresses can be added with ``\and'',
% % just like the second author.
% % To save space, use either the email address or home page, not both
% \and
% Second Author\\
% Institution2\\
% First line of institution2 address\\
% {\tt\small secondauthor@i2.org}
% }

\newcommand*{\affmark}[1][*]{\textsuperscript{#1}}
\newcommand*{\affaddr}[1]{#1}
\newcommand*{\email}[1]{\text{#1}}
\author{%
Chaohui Yu\affmark[1],
~~~Qiang Zhou\affmark[1],
~~~Jingliang Li\affmark[2],
~~~Jianlong Yuan\affmark[1],
~~~Zhibin Wang\affmark[1] \thanks{Corresponding author},
~~~Fan Wang\affmark[1]\\
\affaddr{\affmark[1]Alibaba Group}\\
\affaddr{\affmark[2]University of the Chinese Academy of Sciences}\\
\email{\{huakun.ych,jianchong.zq,gongyuan.yjl,zhibin.waz,fan.w\}@alibaba-inc.com},\\
\email{lijingliang20@mails.ucas.ac.cn}\\
}

\maketitle

%%%%%%%%% ABSTRACT
\begin{abstract}
Modern incremental learning for semantic segmentation methods usually learn new categories based on dense annotations.
%Although promising results are obtained, 
Although achieve promising results, pixel-by-pixel labeling is costly and time-consuming. 
%%任务介绍和挑战
Weakly incremental learning for semantic segmentation (WILSS) is a novel and attractive task, which aims at learning to segment new classes from cheap and widely available image-level labels. 
Despite the comparable results,
%compared to those methods based on dense supervision, 
%the visualizations show that the learned representations of the new classes are less discriminative.
the image-level labels can not provide details to locate each segment, which limits the performance of WILSS.
%%方法
This inspires us to think how to improve and effectively utilize the supervision of new classes given image-level labels while avoiding forgetting old ones.
In this work, we propose a novel and data-efficient framework for WILSS, named \Ours{}.
Specifically, we propose pre-training based co-segmentation to distill the knowledge of complementary foundation models for generating dense pseudo labels. 
%
%We further present a plug-in teacher module optimized by a proposed dense contrastive objective, to more effectively exploit the generated pseudo masks.
We further optimize the noisy pseudo masks with a teacher-student architecture, where a plug-in teacher is optimized with a proposed dense contrastive loss.
Moreover, we introduce memory-based copy-paste augmentation to improve the catastrophic forgetting problem of old classes.
%%实验
Extensive experiments on Pascal VOC and COCO datasets demonstrate the superior performance of our framework, \eg, \Ours{} achieves 70.7\% and 73.3\% in the 15-5 VOC setting, outperforming the state-of-the-art method by 3.4\% and 6.1\%, respectively.

\end{abstract}

%主要贡献是提出了一种针对WILSS的框架
%1.为了在给定image-level label的情况下提升监督信号，我们提出Pre-training Based Co-segmentation利用complementary的pre-trained的大模型生成pseudo mask，我们也是第一个应用大模型到WILSS的工作。

%2.为了生成的mask是有noise的，为了进一步改善pseudo label质量，我们提出Self-training Based Feature Refinement。具体的，我们提出使用一个teacher-student架构，引入一个plug-in的teacher module去在线动态的更新pseudo label，我们首先使用BCE loss去优化pixel的分类能力，同时提出DCL去利用一个batch size里的多类之间的预测信息进一步优化pixel feature，进一步解决noise的 pseudo label问题。

%3.为了缓解灾难性遗忘，我们提出memory-based copy-paste 数据增强策略。

%%%%%%%%% BODY TEXT
%----------------------Introduction--------------------------------------------------------
\section{Introduction}
\label{sec:intro}

\begin{figure}[t!]
\centering
	\includegraphics[width=0.97\linewidth]{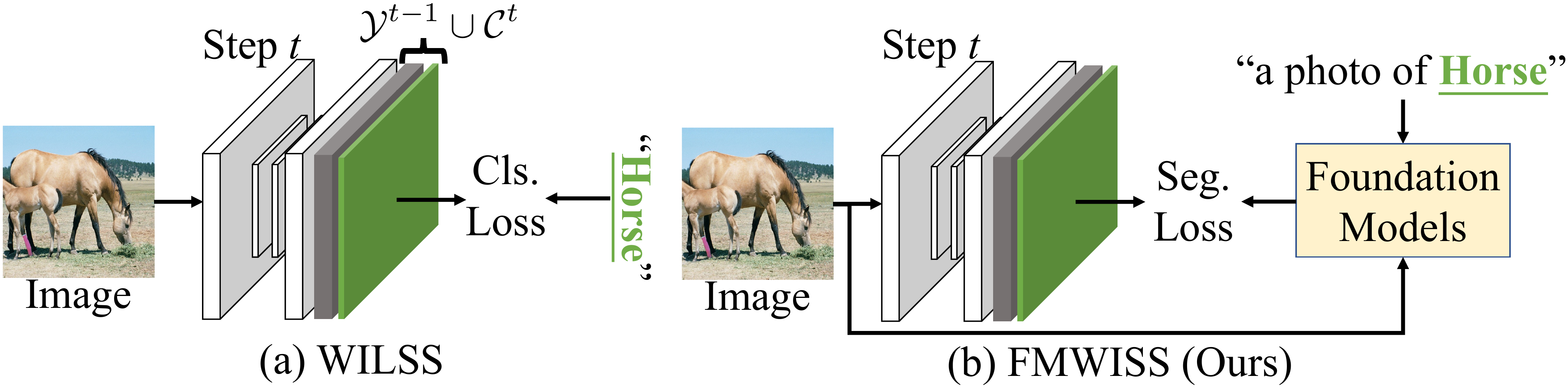}
	\vspace{-.1in}
	\caption{Illustration of the major difference of pipeline between previous WILSS work and \Ours{}. Given a model pre-trained on old classes with pixel-level labels ($\mathcal{Y}^{t-1}$), previous work~\cite{cermelli2022wilson} learn new classes (\eg, \textit{horse}) via image-level labels ($\mathcal{C}^t$), while \Ours{} improves and effectively utilizes the supervision from complementary foundation models.}
	\label{fig:compare_arch}
    \vspace{-.2in}
\end{figure}

%先引入分割，分割的问题，灾难性遗忘
Semantic segmentation is a fundamental task in computer vision and has witnessed great progress using deep learning in the past few years. It aims at assigning each pixel a category label.
Modern supervised semantic segmentation methods~\cite{chen2017deeplab,chen2018encoder} are usually based on published large-scale segmentation datasets with pixel annotations. Despite the promising results, one model pre-trained on one dataset is prone to easily forget learned knowledge when being retrained on another dataset with new classes. 
This phenomenon is known as catastrophic forgetting~\cite{mccloskey1989catastrophic}, which is caused by large changes of model parameters to model new samples with novel categories without accessing old samples.

%再引出一种可用的solution是分割增量学习，大概是什么方法解决的灾难性遗忘
A promising approach to solve such catastrophic forgetting problem is called incremental learning. Many methods have been proposed to solve image classification task~\cite{kirkpatrick2017overcoming,chaudhry2018riemannian,zenke2017continual,li2017learning,dhar2019learning,rebuffi2017icarl,castro2018end,shin2017continual,hou2019learning,wu2018memory,ostapenko2019learning}. 
Recently, a few methods have been presented to address incremental learning for semantic segmentation (ILSS) task, where only new classes of training samples of the current step are labeled with pixel annotations and old classes of the previous step are labeled as background.
Modern ILSS methods can be classified into two categories: regularization-based and replay-based. 
Regularization-based methods~\cite{cermelli2020mib,douillard2021plop,michieli2021sdr} focus on distilling knowledge, \eg, output probability, intermedia features, from pre-trained model of previous step.
Replay-based methods~\cite{maracani2021recall} propose to store the information of previous old classes or web-crawled images and replay for new training steps.
%并说明supervised分割增量学习的缺点，即需要新类的dense标注
However, a key barrier to further develop these methods is the requirement for pixel-level annotations for new classes.
%引出最近WILSON，弱增量学习，优势和缺点
%缺点：image-level label不能提供丰富的信息去定位每个segment
Very recently, WILSON~\cite{cermelli2022wilson} first proposes a new task, weakly incremental learning for semantic segmentation (WILSS), to incrementally update the model from image-level labels for new classes. Despite the comparable results, the image-level labels can not provide details to accurately locate each segment, which limits the performance and development of WILSS.
%the visualization shows that the learned representations of new classes are less discriminative.

%然后给出我们的弱增量方案FMWILSS
In this work, we explore to improve and more effectively utilize the supervision of new classes given image-level labels while preserving the knowledge of old ones. We propose a \underline{\textbf{F}}oundation \underline{\textbf{M}}odel drives \underline{\textbf{W}}eakly \underline{\textbf{I}}ncremental learning for \underline{\textbf{S}}emantic \underline{\textbf{S}}egmentation framework, dubbed FMWISS.

Firstly, as shown in Figure~\ref{fig:compare_arch}, we are the first attempt to leverage pre-trained foundation models to improve the supervision given image-level labels for WILSS in a training-free manner. 
To be specific, we propose pre-training based co-segmentation to distill the knowledge of vision-language pre-training models (\eg, CLIP~\cite{radford2021clip}) and self-supervised pre-training models (\eg, iBOT~\cite{zhou2021ibot}), which can be complementary to each other. However, it is not trivial to apply the pre-trained models.
%We first follow the MaskCLIP~\cite{zhou2021maskclip} mechanism to 
We first adapt CLIP for category-aware dense mask generation. Based on the initial mask for each new class, we then propose to extract compact category-agnostic attention maps with seeds guidance using self-supervised models. We finally refine the pseudo masks via mask fusion. 
We further propose to optimize the still noisy pseudo masks with a teacher-student architecture, where the plug-in teacher is optimized with the proposed dense contrastive loss. Thus we can more effectively utilize the pseudo dense supervision.
%
%We also present a plug-in teacher module (ASPP in Figure~\ref{fig:framework}) to refine and update the pseudo masks.
%Then, we present a dense contrastive loss to more effectively exploit the pseudo masks of new classes by compacting intra-class features and seperating inter-class features.
%
Finally, we present memory-based copy-paste augmentation to remedy the forgetting problem of old classes and can further improve the performance.

%贡献总结
The contributions of this paper are as follows:
\begin{itemize}
\itemsep -0.1cm
\item We present a novel and data-efficient WILSS framework, called FMWISS, which is the first attempt to utilize complementary foundation models to improve and more effectively use the supervision given only image-level labels.

\item We propose pre-training based co-segmentation to generate dense masks by distilling both category-aware and category-agnostic knowledge from pre-trained foundation models, which provides dense supervision against original image labels.

\item To effectively utilize pseudo labels, we use a teacher-student architecture with a proposed dense contrastive loss to dynamically optimize the noisy pseudo labels.

\item We further introduce memory-based copy-paste augmentation to remedy the forgetting problem of old classes and can also improve performance.

\item Extensive experiments on Pascal VOC and COCO datasets demonstrate the significant efficacy of our \Ours{} framework.
\end{itemize}

%----------------------Related Work--------------------------------------------------------
\section{Related Work}

\noindent
\textbf{Incremental Learning for Semantic Segmentation.} In addition to an exhaustive exploration of incremental learning for image classification~\cite{kirkpatrick2017overcoming,chaudhry2018riemannian,zenke2017continual,li2017learning,dhar2019learning,rebuffi2017icarl,castro2018end,shin2017continual,hou2019learning,wu2018memory,ostapenko2019learning}, a relatively few methods~\cite{cermelli2020mib,douillard2021plop,klingner2020cil,maracani2021recall,michieli2019ilt,michieli2021sdr} have been proposed to tackle the incremental learning for semantic segmentation task, which can be classified into regularization-based and replay-based methods.
Regularization-based methods~\cite{cermelli2020mib,douillard2021plop,michieli2021sdr} focus on preserving knowledge from previous training steps. For instance, 
MiB~\cite{cermelli2020mib} proposes a modified version of traditional cross-entropy and knowledge distillation loss terms to regularize the probability of old classes and distill previous knowledge respectively, so as to remedy the background shift problem.
PLOP~\cite{douillard2021plop} proposes a multi-scale pooling technique to preserve long and short-range spatial relationships at the feature level.
SDR~\cite{michieli2021sdr} proposes to optimize the class-conditional features by minimizing feature discrepancy of the same class.
In addition, as a replay-based method, RECALL~\cite{maracani2021recall} uses web-crawled images with pseudo labels to remedy the forgetting problem.
Pixel-by-pixel labeling for semantic segmentation is time-consuming and labor-intensive. Recently, some literature proposes to attain segmentations from cheaper and more available supervisions, \eg, sparse point~\cite{bearman2016s}, and image-level label~\cite{huang2018weakly,lee2019ficklenet,sun2020mining}, which has been attracting more and more attention these years. Most image-based weakly supervised semantic segmentation methods~\cite{araslanov2020ss,lee2021eps,wang2020seam} leverage image-level labels to optimize the class activation map (CAM) and then extract pseudo dense annotations. 
However, the image-level label is rarely explored in incremental learning for semantic segmentation.

Very recently, WILSON~\cite{cermelli2022wilson} first proposes a novel weakly incremental learning for semantic segmentation (WILSS) task, which extends a pre-trained segmentation model using only image-level labels and achieves comparable results.
In this work, inspired by WILSON~\cite{cermelli2022wilson}, we present a \Ours{} framework to improve and effectively utilize the image-level labels by distilling the knowledge of the complementary foundation models.

\begin{figure*}[t!]
	\begin{center}
	\includegraphics[width=0.9\textwidth]{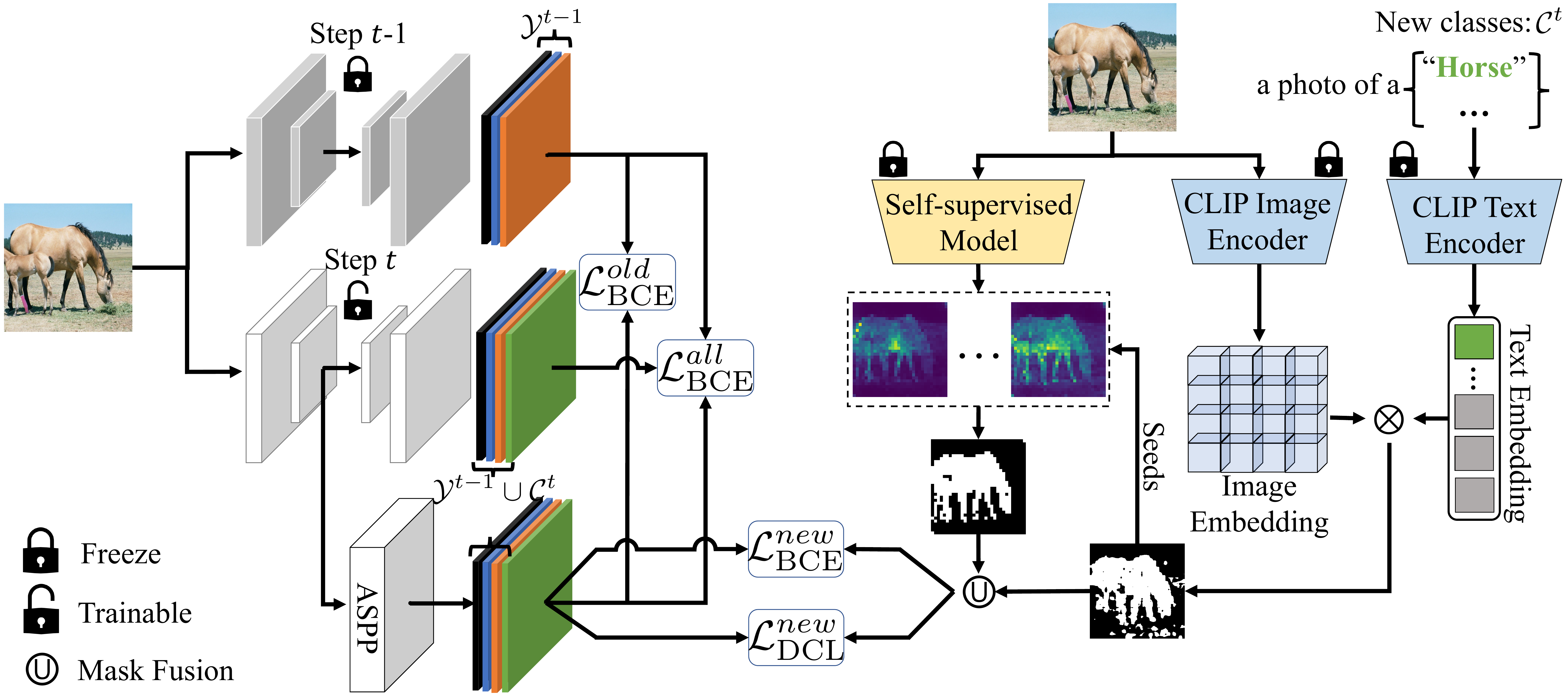}
	\end{center}
    \vspace{-.2in}
	\caption{Illustration of the proposed \Ours{} framework. The plug-in teacher module (ASPP~\cite{chen2014deeplabv2}) is to learn the segments of both old and new classes during training, which is eliminated during inference. The model at step \textit{t} is optimized using the outputs of the pre-trained model at step \textit{t}-1 and the learned logits of the online teacher module.}
	\label{fig:framework}
	\vspace{-.1in}
\end{figure*}

%\paragraph{Visual Foundation Models}
\noindent
\textbf{Visual Foundation Models.} We mainly focus on two kinds of foundation models in computer vision, including the vision-language pre-training (VLP) models and the self-supervised pre-training models.
VLP~\cite{radford2021clip,jia2021scalingalign} plays an important role in multimodal research, \eg, VQA~\cite{antol2015vqa}, text-to-image generation~\cite{ramesh2022hierarchicaldalle2}, zero-shot classification~\cite{zhou2021coop,zhou2022cocoop}. A representative VLP work is CLIP~\cite{radford2021clip}, which jointly trains the image and text encoders on 400 million image-text pairs collected from the web and demonstrates promising results on zero-shot image classification tasks. Recently, MaskCLIP~\cite{zhou2021maskclip} adapts CLIP to zero-shot semantic segmentation in a training-free manner, which illustrates the potential of CLIP in category-aware dense prediction.
Self-supervised visual pre-training can be classified into three categories: contrastive learning based~\cite{oord2018infonce,chen2020simple,he2020momentum}, distillation based~\cite{grill2020bootstrap,caron2021dino}, and masked image modeling based~\cite{he2022mae,zhou2021ibot}. Among these methods, iBOT~\cite{zhou2021ibot} and DINO~\cite{caron2021dino} are two representative approaches to automatically perform class-agnostic dense features modeling.
These two kinds of foundation models can be complementary to each other.

%------------------------------------Method-----------------------------------------

\section{Method}

\subsection{Problem Definition and Notation}
Let $\mathcal{X}$ be the input image space and each image $x \in \mathcal{X}$ consists of a set of pixels $\mathcal{I}$ with $|\mathcal{I}| = H \times W$. Let $\mathcal{Y}$ be the label space.
In the incremental learning for semantic segmentation setting~\cite{cermelli2020mib}, the training procedure is arranged into multiple steps, and each learning step $t$ will involve novel classes $\mathcal{C}^t$ with pixel annotations, constructing a new label set $\mathcal{Y}^t = \mathcal{Y}^{t-1} \cup \mathcal{C}^t$.
However, different from the original incremental setting, in the novel weakly incremental learning for semantic segmentation (WILSS) setting, recently proposed by WILSON~\cite{cermelli2022wilson}, the pixel annotations are only provided for the initial step, \ie, $t=0$. For the following steps, we can only access to training sets with image-level annotations for new classes, and can not access to the training samples of previous training steps anymore. The goal is to learn and update a model to perform segmentation on new classes without forgetting old classes.

%\subsection{Seeds-guided Co-segmentation}
\subsection{Pre-training Based Co-segmentation}

It is still challenging to use only image-level labels to supervise the dense prediction tasks, \eg, semantic segmentation, since image-level labels can not provide detailed information to accurately locate each segment. This limitation inspires us to investigate the following question:  \textit{how to improve the supervision of new classes from image-level labels?}
To tackle this question, we propose a pre-training based co-segmentation method to utilize the knowledge of foundation models in a training-free manner. We distill the complementary knowledge of two kinds of foundation models, including the vision-language pre-training models, \eg, CLIP~\cite{radford2021clip}, and the self-supervised pre-training models, \eg, iBOT~\cite{zhou2021ibot}, DINO~\cite{caron2021dino}.

% \noindent
% \textcolor{red}{to remove} \textbf{Review of CLIP.} CLIP~\cite{radford2021clip} trains a pair of image and text encoders on 400 million web-scale image-text pairs via a contrastive loss, which demonstrates a promising potential to explicitly leverage language for solving computer vision tasks efficiently. After pre-training, the knowledge of CLIP can be widely used in zero-shot classification tasks~\cite{zhou2021learning,zhang2021tipadapter}. Let $x$ be the input image of image encoder $f_I$ and $t_y$ be the input prompt of text encoder $f_T$, \eg, ``{\tt a photo of a [CLASS]}'' where ``{\tt [CLASS]}'' is replaced with the $y$-th class name. The predicted classification probability is calculated as:
% \begin{equation}
% p(y|x) = \frac{\text{exp}(\text{cos}(f_I(x), f_T(t_y)) / \tau)} {\sum^{K}_{i=1} \text{exp}(\text{cos}(f_I(x), f_T(t_i))/\tau)},
% \label{eq:clip_cls}
% \end{equation}
% where $\text{cos}(\cdot,\cdot)$ denotes the cosine similarity, $\tau$ is a learned temperature parameter, and $K$ is the total categories of a given dataset.

\noindent
\textbf{Initial Mask.} 
%To get dense prediction of a new class image, we draw inspiration from the recent MaskCLIP~\cite{zhou2021maskclip}, which indicates that the dense visual feature of CLIP image encoder can be correlated with a given text embedding of CLIP text encoder. 
We believe that the pre-trained vision-language model, \eg, CLIP, has encoded rich semantic information in its features as it learns to associate image with language caption from 400 million image-text pairs. To get dense prediction of a new class image, we apply the pre-trained CLIP model to extract category-aware pixel annotations given image-level labels. 
As shown in Figure~\ref{fig:framework}, to be specific, given an image $x$ of step $t$ with image-level labels $\mathcal{C}^t$, we first extract dense image features $F$ from the CLIP image encoder $f_I$. Then, project  $F$ via the final linear projection layer $f_{proj}$ of $f_I$.
\begin{equation}
\overline{F} = | f_{proj}(F) |_{2}, ~~ \overline{F} \in \mathbb{R}^{h \times w \times d},
\end{equation}
where $|\cdot|_{2}$ denotes L2 normalization along the channel dimension, $d$ is the feature dimension of the joint space.
We then compute the text embeddings by taking as input the prompt with target classes $\mathcal{C}^t$ of step $t$:
\begin{equation}
\overline{T} = | f_T({\tt prompt} (\mathcal{C}^t)) |_2, ~~ \overline{T} \in \mathbb{R}^{\mathcal{C}^t \times d},
\end{equation}
where $f_T$ is the CLIP text encoder, ${\tt prompt}(\cdot)$ denotes prompt engineering, which ensembles multiple prompt templates as in~\cite{radford2021clip}.

We then compute the pixel-text score maps using the language-compatible image feature embeddings $\overline{F}$ and the text embeddings $\overline{T}$ by:
\begin{equation}
\mathcal{M}_{init} = \overline{F} \cdot \overline{T}^{\top}, ~\mathcal{M}_{init} \in \mathbb{R}^{h \times w \times \mathcal{C}^t},
\end{equation}
which indicates that each pixel will be assign a score for each class in $\mathcal{C}^t$, and $\mathcal{M}_{init}$ can be viewed as the initial segmentation results with category information.

%The resulting text embeddings of each class is used as the weights of a $1 \times 1$ convolution classifier. We can thus get the initial mask:
%\begin{equation}
%\mathcal{M}_{init} = \mathop{\arg\max}_{C^t}~{\tt F.conv2d}(\overline{F}^{new}_{i}, ~\overline{T}^{new}),  %~\mathcal{M}_{init} \in \mathbb{R}^{h \times w \times C^t},
%\end{equation}
%where $\mathcal{M}_{init} \in \mathbb{R}^{h \times w \times C^t}$ and ${\tt F.conv2d}$ denote the training-free convolution operation.

%iBOT, 引出seeds-guided, 给一个init mask -》 seeds -〉 co-seg图
\noindent
\textbf{Refine Mask via Seeds Guidance.} The pseudo mask $\mathcal{M}_{init}$ generated by CLIP~\cite{radford2021clip} can provide rich category-aware pixel annotations, but the mask is noisy since the training paradigm of CLIP based on image-text pairs doomed to be good at instance-level classification rather than segmentation.
%The pseudo mask generated by MaskCLIP~\cite{zhou2021maskclip} can not guarantee a good highlight of a target new classes, since MaskCLIP is a training-free adaptation of CLIP for dense prediction. However, the training paradigm of CLIP based on image-text pairs doomed to be good at instance-level classification rather than segmentation.
%
To improve the mask quality, we propose to distill the knowledge of another kind of foundation models, \ie, self-supervised pre-training models, which have shown promising performance in local feature modeling~\cite{caron2021dino,zhou2021ibot}. These models can produce compact category-agnostic attention maps. However, \textit{how to extract segmentations for a target class given an image that may contain multiple objects?}
To address this issue, we propose to refine the initial mask via category-specific seeds guidance. 
%considering that MaskCLIP can generate category-aware but coarse segmentations and iBOT can generate category-agnostic but more compact attentions, 
Specifically, we randomly select $N=\{(i_p, j_p)\}^{N}_{p=1}$ seed points from initial mask $\mathcal{M}_{init} \in \mathbb{R}^{h \times w \times \mathcal{C}^t}$ for each target class $c \in \mathcal{C}^t$, and extract the corresponding attention maps from the pre-trained self-supervised model.
Let $S$ denotes the image encoder of the self-supervised model, $S(x) \in \mathbb{R}^{h \times w \times n}$ denotes the output attention maps of the last self-attention block. We extract the category-aware attention map with the guidance of seeds as follows. For simplicity, we only show the calculation on one class $c \in \mathcal{C}^t$:
\begin{equation}
\mathcal{M}^{c}_{seeds} = \big [\frac{1}{N} \sum^{N}_{p=1}  \frac{1}{n} \sum^{n} S(x) \big ]_{\tt binary}, %\mathcal{M}^{c}_{refine} \in \mathbb{R}^{h \times w},
\label{eq:m_seeds}
\end{equation}
where $n$ denotes the number of attention heads. The $N$ seed points are randomly sampled from the foreground of binarized $\mathcal{M}_{init}$ for each new class and training step.
$[\cdot]_{\tt binary}$ is a binarization operation that sets all the values greater than the threshold $\mathcal{T}$ to 1, otherwise 0. $\mathcal{T}$ is dynamically updated to keep the top $\mathcal{K}$\% ($\mathcal{K}=70$ by default) locations from the averaged attention map.
As shown in Figure~\ref{fig:seeds}, we visualize extracted attention maps of two classes (\textit{horse, dog}), nine seeds ($N=9$) can already show good clustering performance.

Finally, we get the refined mask via simple mask fusion for better performance:
\begin{equation}
\mathcal{M}_{refine} = \mathcal{M}_{init} \cup \mathcal{M}_{seeds}, ~~\mathcal{M}_{refine} \in \mathbb{R}^{h \times w \times \mathcal{C}^t},
\label{eq:fusion}
\end{equation}
where $\cup$ represents mask fusion operation, which is the union operation in our experiments by default.

%然后提出ASPP module动态更新这个label，作为我们的baseline
%Inspired by previous WILSS work~\cite{cermelli2022wilson} that use two convolution layers to learn Class Activation Map (CAM) from image-level labels, we introduce a plug-in teacher module (ASPP network in Figure~\ref{fig:framework}) to dynamically learn better pseudo masks during training.

%\subsection{Dense Contrastive Learning}
\subsection{Pseudo Label Optimization}
%解决如何更好的利用生成的pseudo mask

\begin{figure}[t!]
\centering
	\includegraphics[width=0.95\linewidth]{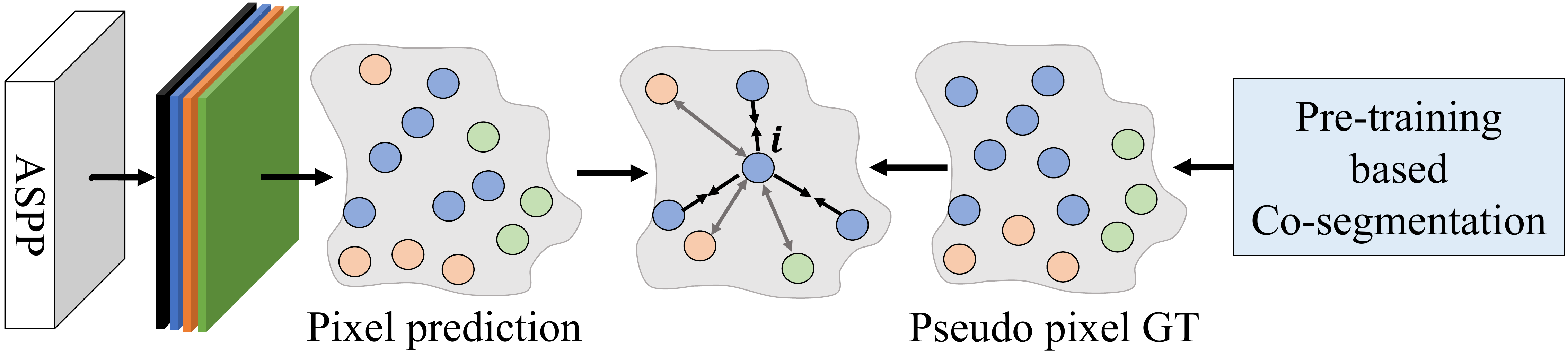}
    \vspace{-.1in}
	\caption{Illustration of the dense contrastive loss calculation process. The colorful points represent pixels with different categories.}
	\label{fig:contrast}
	\vspace{-.1in}
\end{figure}

%Inspired by previous WILSS work~\cite{cermelli2022wilson} that use two convolution layers to learn Class Activation Map (CAM) from image-level labels, we introduce a plug-in teacher module (ASPP network in Figure~\ref{fig:framework}) to dynamically learn better pseudo masks during training.

Previous WILSS literature~\cite{cermelli2022wilson} use learned CAM to supervise the segmentation learning of new classes, and CAM is optimized with a binary cross-entropy (BCE) loss against one-hot image-level labels.
Now, we have the generated pseudo pixel labels that can provide more information than image-level labels, a natural question is: \textit{how to effectively utilize such supervision?}

We propose to use a teacher-student architecture to further optimize the still noisy pseudo mask. To be specific, by taking the segmentation model as student model, we introduce a plug-in teacher module (ASPP network in Figure~\ref{fig:framework}) to dynamically learn better pseudo masks during training.
To learn the teacher module, we first propose to use the pixel-wise binary cross-entropy loss $\mathcal{L}^{new}_{\text{BCE}}$ to supervise the predictions of the new classes at step $t$ as follows, and we leave the old classes optimization in the next section. 
\begin{equation}
\begin{split}
\mathcal{L}^{new}_{\text{BCE}} = - \frac{1}{|\mathcal{C}^{t}| |\mathcal{I}|} \sum_{i \in \mathcal{I}} \sum_{c \in \mathcal{C}^t} \mathcal{M}^{c, i}_{refine}\log(p^{c, i}) \\
+ (1 - \mathcal{M}^{c, i})\log(1 - p^{c, i}),
 \end{split}
 \label{eq:pixelbce}
\end{equation}
where $p^{c, i}$ denotes the predicted probability on new class $c$ of pixel $i$.
However, the pixel-wise BCE loss mainly focus on the optimization of the target foreground class $c$ of current input image and treat all other classes as background, which ignores the correlation among pixels. To better utilize the multi-class predictions and corresponding pixel-wise pseudo labels among the entire dataset, inspired by the InfoNCE~\cite{oord2018infonce} loss in unsupervised representation learning, we propose to perform dense contrastive learning.

Specifically, as depicted in Figure~\ref{fig:contrast}, for a pixel $i$ of a new class image and its corresponding pseudo annotation, we collect all the pixels with the same class label as $i$ to compose positive samples $\mathbf{P}_i$, and collect the points of other classes to compose negative samples $\mathbf{N}_i$. Formally, our dense contrastive loss $\mathcal{L}^{new}_{\text{DCL}}$ can be calculated as follows. For simplicity, we only show the loss on pixel $i$:
\begin{equation}
\begin{aligned}
& \mathcal{L}^{new}_{\text{DCL}} \\
& = \frac{1}{|\mathbf{P}_i|} \sum_{q_{+} \in \mathbf{P}_i}-\log \frac{\text{exp}(i \cdot q_{+}/\tau)} {\text{exp}(i \cdot q_{+}/\tau) + \sum\limits_{q_{-} \in \mathbf{N}_i} \text{exp}(i \cdot q_{-}/\tau)},
\end{aligned}
\label{eq:dcl}
\end{equation}
where $\tau$ is a temperature term (0.1 by default). For training efficiency, we randomly sample only ten points for each contained class of the current mini-batch.

The pixel-wise BCE loss in Eq.~(\ref{eq:pixelbce}) and the dense contrastive loss in Eq.~(\ref{eq:dcl}) can be complementary to each other, which help the teacher module to learn discriminative pixel features as well as regularize the pixel feature space by intra-class and inter-class pixel feature modeling.

\subsection{Memory-based Copy-Paste Augmentation}

\begin{figure}[t!]
\centering
	\includegraphics[width=0.8\linewidth]{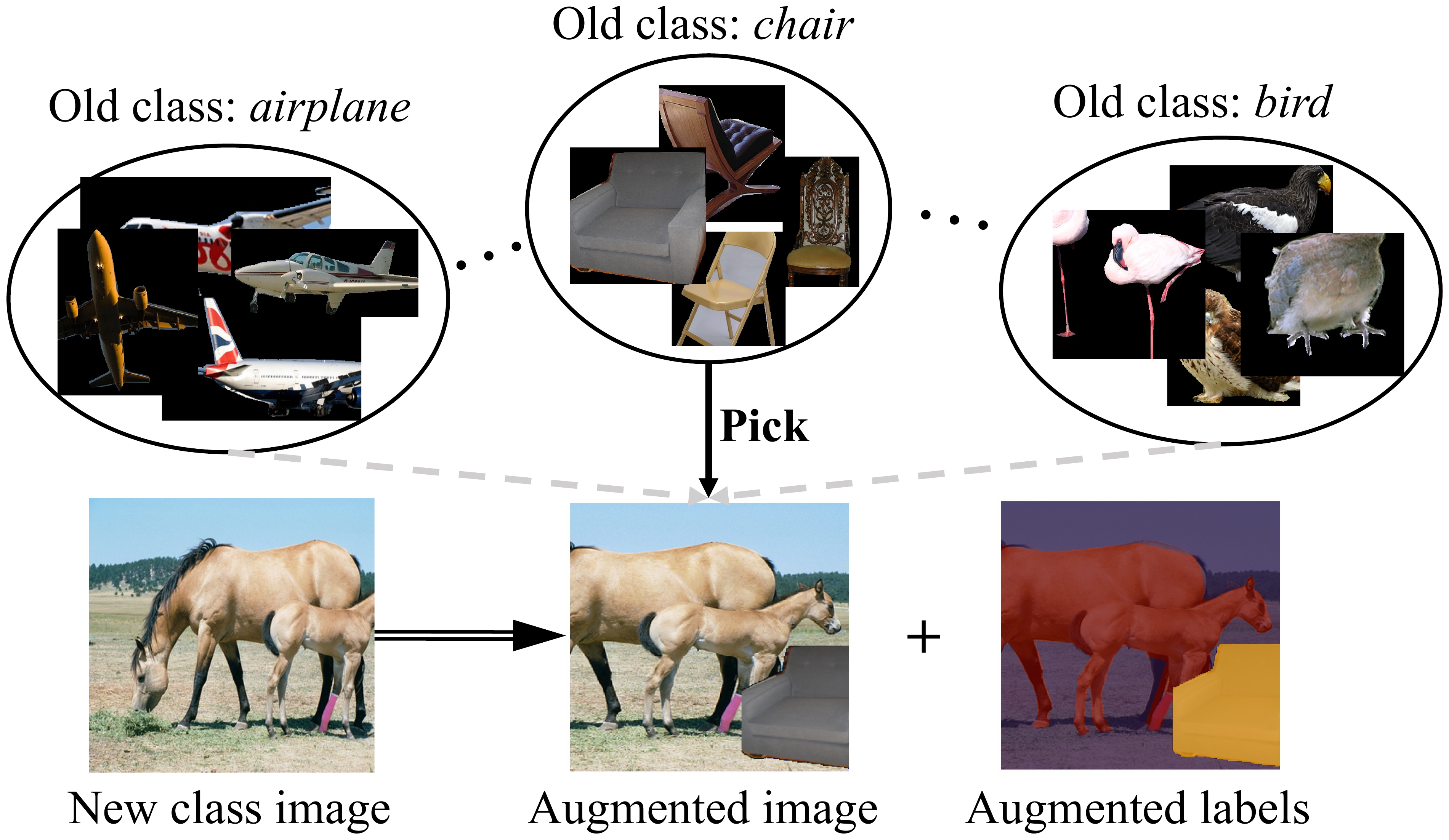}
    \vspace{-.1in}
	\caption{Illustration of the memory-based copy-paste augmentation. Notably, we will not use the ground-truth labels of new classes of augmented labels.}
	\label{fig:copypaste}
	\vspace{-.1in}
\end{figure}

In addition to improving and effectively leveraging the supervision of new classes for WILSS, we propose a memory-based copy-paste augmentation strategy to stabilize the learning of old classes and can further improve the performance of the segmentation model.
As shown in Figure~\ref{fig:copypaste}, we first construct a memory bank for each old class, and each class archive will store $\mathcal{B}$ foreground instances and segmentation labels during the base model training. Then, in step $t$, we randomly pick one pair of foreground images and labels from a randomly selected old class archive, and randomly paste them to the new class image. 
Now, the training samples contain new class images at step $t$ as well as old class images and pixel labels at step $t$-1. We thus optimize the old class learning of the teacher module as:
\begin{equation}
\begin{split}
\mathcal{L}^{old}_{\text{BCE}} = - \frac{1}{|\mathcal{Y}^{t-1}| |\mathcal{I}|} \sum_{i \in \mathcal{I}} \sum_{c \in \mathcal{Y}^{t-1}} g(f^{t-1}_{\theta}(\hat{x}))^{c,i}\log(p^{c, i}) \\
 + (1 - g(f^{t-1}_{\theta}(\hat{x}))^{c,i})\log(1 - p^{c, i}),
 \end{split}
\end{equation}
\begin{equation}
g(f^{t-1}_{\theta}(\hat{x}))^{c,i} = 
\begin{cases}
1, & \hat{x}^{c,i}~\text{augmented}, \\ 
g(f^{t-1}_{\theta}(\hat{x}))^{c,i}, & \text{otherwise},
\end{cases} 
\end{equation}
where $g(\cdot)$ is the logistic function, $f^{t-1}_{\theta}$ is the trained model at step $t$-1, $p^{c, i}$ is the predicted probability on old class $c$ of pixel $i$, $\hat{x}$ denotes the augmented image.

\subsection{Overall Optimization}
%上面让teacher module学好了old+new，下面用teacher 蒸馏 model at step t
We optimize the segmentation model $f_{\theta}^{t}$ at step $t$ by distilling the knowledge of the trained model $f^{t-1}_{\theta}$ and the dynamically updated teacher module $T^{t}$. 
Since $T^{t}$ is optimized mainly through the binary cross-entropy loss, we use the BCE loss to distill the prediction of $T^{t}$ to model $f_{\theta}^{t}$.
Considering that the learned pseudo mask is not perfect, we use the soft pixel labels as the final supervision for new classes $\mathcal{C}^t$, and use the weighted average value of the old model and teacher module outputs as the supervision for old classes:
\begin{equation}
q^{c,i} = 
\begin{cases}
\alpha |T^{t}(x)|_{\text{hard}} + (1-\alpha) g(T^{t}(x))  , & \text{if}~ c \in \mathcal{C}^{t}, \\ 
\beta g(f_{\theta}^{t-1}(x)) + (1-\beta) g(T^{t}(x)), & \text{otherwise},
\end{cases}
\label{eq:qci}
\end{equation}
where $|\cdot|_{\text{hard}}$ denotes one-hot operation to set one to the class with the maximum score for each pixel and zero to others. $\alpha$ and $\beta$ are trade-off parameters and we set $\alpha=0.5, \beta=0.9$ by default. Then, the BCE loss for $f_{\theta}^{t}$ is:
\begin{equation}
\begin{split}
\mathcal{L}^{all}_{\text{BCE}} = - \frac{1}{|\mathcal{Y}^{t}||\mathcal{I}|} \sum_{i \in \mathcal{I}} \sum_{c \in \mathcal{Y}^{t}} q^{c,i}\log(p^{c, i}) \\
+ (1 - q^{c,i})\log(1 - p^{c, i}),
\end{split}
\end{equation}
where $\mathcal{Y}^{t}$ is the set of all seen classes and $p=f_{\theta}^{t}(x)$ represents the output of segmentation model at step $t$.

The overall learning objective is as follows:
\begin{equation}
\mathcal{L}_{overall} =  \mathcal{L}^{new}_{\text{BCE}} + \lambda \mathcal{L}^{new}_{\text{DCL}} + \mathcal{L}^{old}_{\text{BCE}} + \mathcal{L}^{all}_{\text{BCE}},
\end{equation}
where $\lambda$ is the loss weight of $\mathcal{L}^{new}_{\text{DCL}}$.

%----------------------Experiments--------------------------------------------------------
\section{Experiments}

\subsection{Datasets and Protocols}
To ensure a fair comparison, we follow the experimental settings of the state-of-the-art (SoTA) WILSS method WILSON~\cite{cermelli2022wilson} for datasets and protocols. Different from~\cite{cermelli2020mib,maracani2021recall} rely on pixel-wise annotation on new classes, we only use image-level labels for novel classes as WILSON.

\noindent
\textbf{Datasets:} we consider two standard evaluation benchmarks including Pascal VOC 2012~\cite{everingham2010pascalvoc} and COCO~\cite{lin2014coco}. COCO consists of 118,287 and 5,000 images for training and validation with 80 annotated object classes. Pascal VOC is composed of 1,464 training and 1,449 validation images with 20 labeled object classes. We follow the standard method~\cite{ahn2018learning,kolesnikov2016seed,cermelli2022wilson} to augment the VOC dataset with images from~\cite{hariharan2011semantic}, building up 10,582 and 1,449 images for training and validation, respectively. 
We also follow the practice of~\cite{cermelli2022wilson} to use the train split and the annotation of COCO-Stuff~\cite{caesar2018cocostuff} that addresses the annotation overlapping problem of COCO~\cite{lin2014coco}.

\noindent
\textbf{Protocols:} previous works~\cite{cermelli2020mib} introduce two different incremental learning protocols: \textit{disjoint} and \textit{overlap}. 
In \textit{disjoint}, images of each training step only contain pixels of previous seen and current classes.
In \textit{overlap}, each training step contains all the images, where pixels can belong to any class. Thus, the overlap protocol is more realistic and challenging compared to the disjoint. In our experiments, we follow the previous WILSS work~\cite{cermelli2022wilson} to apply these two protocols on the VOC dataset, including \textbf{15-5 VOC}, where 15 base classes are learned in the first training step and 5 new classes are continuously learned in the second step; \textbf{10-10 VOC}, where 10 base classes are learned in the first step and another 10 new classes are added in the second step.
In addition, we also verify our method on the \textbf{COCO-to-VOC} protocol, which is a new incremental learning scenario proposed in~\cite{cermelli2022wilson}. To be specific, in the first step, we learn the 60 classes of COCO that do not appear in the VOC dataset. Then, we continuously learn the 20 classes of VOC. Following common practice~\cite{cermelli2020mib,maracani2021recall,cermelli2022wilson}, we report the standard mean Intersection over Union (mIoU) results on the validation sets.

\subsection{Implementation Details}
As in WILSON~\cite{cermelli2022wilson}, we use a Deeplab V3~\cite{chen2017deeplabv3} with a ResNet-101~\cite{he2016deep} backbone for VOC and a Wide-ResNet-38 for COCO, both pre-trained on ImageNet~\cite{deng2009imagenet}.
%As in~\cite{cermelli2020mib,cermelli2022wilson}, in-place activated BN is used to reduce runtime memory. 
We train all the models for 40 epochs with a batch size of 24 and the SGD optimizer with a learning rate of 1$e^{-3}$, momentum of 0.9, weight decay of 1$e^{-4}$. Before training the segmentation model, we first warm up the teacher module for five epochs. We set $\lambda=0.1$, $\alpha=0.5$, and $\beta=0.9$.
As for foundation models, in terms of the VLP model, we find that MaskCLIP~\cite{zhou2021maskclip} is an alternative solution to adapt CLIP for better dense prediction, and we use its mechanism based on the pre-trained CLIP with ViT-B architecture. 
In terms of the self-supervised model, we use pre-trained iBOT~\cite{zhou2021ibot} with ViT-L architecture by default.

\subsection{Baselines}
Since weakly incremental learning for semantic segmentation (WILSS) is a novel setting proposed by WILSON~\cite{cermelli2022wilson}, we also compare our framework with both supervised incremental learning and weakly supervised semantic segmentation methods as in~\cite{cermelli2022wilson}.
For supervised incremental learning methods using dense pixel-wise annotations, we compare with eight representative state-of-the-art works, including LWF~\cite{li2017lwf}, LWF-MC~\cite{rebuffi2017lwfmc}, ILT~\cite{michieli2019ilt}, MiB~\cite{cermelli2020mib}, PLOP~\cite{douillard2021plop}, CIL~\cite{klingner2020cil}, SDR~\cite{michieli2021sdr}, and RECALL~\cite{maracani2021recall}.
As for the weakly supervised semantic segmentation methods adopted to the incremental learning scenario, we compare with the reported results with the pseudo labels generated from class activation maps (CAM), SEAM~\cite{wang2020seam}, SS~\cite{araslanov2020ss}, and EPS~\cite{lee2021eps} as in~\cite{cermelli2022wilson}.

\subsection{Results.}
\subsubsection{Performance on 15-5 VOC}
In this setting, for a fair comparison, we also introduce 5 classes in the incremental step as in~\cite{cermelli2022wilson}: \textit{plant, sheep, sofa, train, tv monitor.}
As shown in Table~\ref{tbl:voc15-5}, our \Ours{} achieves the new state-of-the-art results in all the settings (disjoint and overlap) compared to all the pixel label based and image label based methods.
To be more specific, compared to pixel label based methods, our WILSS method \Ours{} even outperforms the best method by 3.5\% and 3.2\% in the two settings, respectively.
Compared to image label based methods, \Ours{} improves the overall performance by \textbf{3.4\%} and \textbf{6.1\%} in the two settings against previous SoTA WILSON~\cite{cermelli2022wilson}.
Notably, \Ours{} significantly improves the performance on new classes (16-20) by \textbf{7.0\%} and \textbf{12.8\%} in the disjoint and overlap settings, respectively.

\def\pixel{{P}}
\def\image{{I}}
\begin{table}[t!]
\centering
\resizebox{\linewidth}{!}{%
\begin{tabular}{rc|lll|lll}
\toprule
\multirow{2}{*}{Method} & \multirow{2}{*}{Sup}      & \multicolumn{3}{c|}{Disjoint} & \multicolumn{3}{c}{Overlap} \\
                        &                           & 1-15    & 16-20    & All     & 1-15    & 16-20    & All    \\ 
\midrule
Joint$^{\ast}$                      				& \pixel{}         & 75.5    & 73.5     & 75.4    & 75.5    & 73.5     & 75.4   \\ \hline
FT$^{\ast}$                         				& \pixel{}         & 8.4     & 33.5     & 14.4    & 12.5    & 36.9     & 18.3  \\
LWF$^{\ast}$~\cite{li2017lwf}       			& \pixel{}         & 39.7    & 33.3     & 38.2    & 67.0    & 41.8     & 61.0   \\
LWF-MC$^{\ast}$~\cite{rebuffi2017lwfmc}  	& \pixel{}    	   & 41.5    & 25.4     & 37.6    & 59.8    & 22.6     & 51.0   \\
ILT$^{\ast}$~\cite{michieli2019ilt}      			& \pixel{}       & 31.5    & 25.1     & 30.0    & 69.0    & 46.4     & 63.6   \\
CIL$^{\ast}$~\cite{klingner2020cil}      		& \pixel{}       & 42.6    & 35.0     & 40.8    & 14.9    & 37.3     & 20.2   \\
MIB$^{\ast}$~\cite{cermelli2020mib}      		& \pixel{}        & 71.8    & 43.3     & 64.7    & 75.5    & 49.4     & 69.0   \\
PLOP~\cite{douillard2021plop}            		& \pixel{}       & 71.0    & 42.8     & 64.3    & \underline{75.7}    & 51.7     & \underline{70.1}   \\
SDR$^{\ast}$~\cite{michieli2021sdr}     		& \pixel{}        & \underline{73.5}    & 47.3     & \underline{67.2}    & 75.4    & 52.6     & 69.9   \\
RECALL$^{\ast}$~\cite{maracani2021recall}  	& \pixel{}       & 69.2    & \underline{52.9}     & 66.3    & 67.7    & \underline{54.3}     & 65.6   \\ 
\midrule
CAM$^{\dagger}$                     						& \image{}   & 69.3    & 26.1     & 59.4    & 69.9    & 25.6     & 59.7   \\
SEAM$^{\dagger}$~\cite{wang2020seam}        			& \image{}    & 71.0    & 33.1     & 62.7    & 68.3    & 31.8     & 60.4   \\
SS$^{\dagger}$~\cite{araslanov2020ss}       				& \image{}   & 71.6    & 26.0     & 61.5    & 72.2    & 27.5     & 62.1   \\
EPS$^{\dagger}$~\cite{lee2021eps}           				& \image{}    & 72.4    & 38.5     & 65.2    & 69.4    & 34.5     & 62.1   \\
WILSON$^{\dagger}$~\cite{cermelli2022wilson}  			& \image{}    & 73.6    & 43.8     & 67.3    & 74.2    & 41.7     & 67.2   \\ %\hline
%\Ours{} (Ours)                  					& Image    & \textbf{75.6 (+2.0)}  & \textbf{50.8 (+7.0)}   & \textbf{70.5 (+3.2)}  & \textbf{78.0 (+3.8)}  & \textbf{54.2 (+12.5)}   & \textbf{73.0 (+5.8)} \\
\Ours{} (Ours)                              & \image{}    & \textbf{75.9 (+2.3)}  & \textbf{50.8 (+7.0)}  & \textbf{70.7 (+3.4)}  & \textbf{78.4 (+4.2)}  & \textbf{54.5 (+12.8)} & \textbf{73.3 (+6.1)} \\
\bottomrule
\end{tabular}%
}
\vspace{-.1in}
\caption{Results on the 15-5 setting of Pascal VOC. $^{\ast}$ denotes the results are from~\cite{maracani2021recall}, $^{\dagger}$ denotes the results are from~\cite{cermelli2022wilson}. `P' and `I' indicate pixel-level and image-level labels, respectively. The best methods of using image-level labels and using pixel-level labels as supervision are \textbf{bold} and \underline{underlined}, respectively.}
\label{tbl:voc15-5}
\vspace{-.1in}
\end{table}

\subsubsection{Performance on 10-10 VOC}
In this setting, for a fair comparison, we also introduce 10 classes in the incremental step as in~\cite{cermelli2022wilson}: \textit{dining table, dog, horse, motorbike, person, plant, sheep, sofa, train, tv monitor.}
As reported in Table~\ref{tbl:voc10-10}, our \Ours{} achieves the new state-of-the-art performance against the image label based methods and achieves \textit{on par} or even better performance than the pixel label based methods.
Specifically, compared to pixel label based methods, we achieve an overall performance of 64.6\% on disjoint protocol, which is very close to ILT's 64.7\%. On overlap protocol, we achieve an overall performance of 69.1\%, which is even 1.8\% higher than SDR's 67.4\%.
Compared to image label based methods, our \Ours{} achieves the best results on all the protocols, and we achieve overall performance improvements of \textbf{3.8\%} and \textbf{4.1\%} in both settings compared to WILSON~\cite{cermelli2022wilson}.

\begin{table}[t!]
\centering
\resizebox{\linewidth}{!}{%
\begin{tabular}{rc|lll|lll}
\toprule
\multirow{2}{*}{Method} & \multirow{2}{*}{Sup} & \multicolumn{3}{c|}{Disjoint} & \multicolumn{3}{c}{Overlap} \\
                        &                      & 1-10    & 11-20    & All     & 1-10    & 11-20    & All    \\ 
\midrule
Joint$^{\ast}$                   					& \pixel{}                & 76.6    & 74.0     & 75.4    & 76.6    & 74.0     & 75.4   \\ \hline
FT$^{\ast}$                        				& \pixel{}                & 7.7     & 60.8     & 33.0    & 7.8     & 58.9     & 32.1  \\
LWF$^{\ast}$~\cite{li2017lwf}                     	& \pixel{}                & 63.1    & 61.1     & 62.2    & \underline{70.7}    & 63.4     & 67.2   \\
LWF-MC$^{\ast}$~\cite{rebuffi2017lwfmc}         & \pixel{}                & 52.4    & 42.5     & 47.7    & 53.9    & 43.0     & 48.7   \\
ILT$^{\ast}$~\cite{michieli2019ilt}                       & \pixel{}                & \underline{67.7}    & \underline{61.3}     & \underline{64.7}    & 70.3    & 61.9     & 66.3   \\
CIL$^{\ast}$~\cite{klingner2020cil}                     & \pixel{}                & 37.4    & 60.6     & 48.8    & 38.4    & 60.0     & 48.7   \\
MIB$^{\ast}$~\cite{cermelli2020mib}                  & \pixel{}                & 66.9    & 57.5     & 62.4    & 70.4    & 63.7     & 67.2   \\
PLOP~\cite{douillard2021plop}                        	& \pixel{}                & 63.7    & 60.2     & 63.4    & 69.6    & 62.2     & 67.1   \\
SDR$^{\ast}$~\cite{michieli2021sdr}     		& \pixel{}                & 67.5    & 57.9     & 62.9    & 70.5    & \underline{63.9}     & \underline{67.4}   \\
RECALL$^{\ast}$~\cite{maracani2021recall}     & \pixel{}                & 64.1    & 56.9     & 61.9    & 66.0    & 58.8     & 63.7   \\ 
\midrule
CAM$^{\dagger}$                     & \image{}                			& 65.4    & 41.3     & 54.5    & 70.8    & 44.2     & 58.5   \\
SEAM$^{\dagger}$~\cite{wang2020seam}                    		& \image{}                & 65.1    & 53.5     & 60.6    & 67.5    & 55.4     & 62.7   \\
SS$^{\dagger}$~\cite{araslanov2020ss}                      		& \image{}                & 60.7    & 25.7     & 45.0    & 69.6    & 32.8     & 52.5   \\
EPS$^{\dagger}$~\cite{lee2021eps}                     			& \image{}                & 64.2    & 54.1     & 60.6    & 69.0    & 57.0     & 64.3   \\
WILSON$^{\dagger}$~\cite{cermelli2022wilson}                  	& \image{}                & 64.5    	& 54.3     		& 60.8    	& 70.4    	& 57.1     & 65.0   \\ %\hline
%\Ours{} (Ours)                  				& Image                &  68.51       &  58.20        	& 64.64     & 73.79    & 62.26   & 69.09 	\\
\Ours{} (Ours)                  					& \image{}                &  \textbf{68.5 (+4.0)}       &  \textbf{58.2 (+3.9)}        	& \textbf{64.6 (+3.8)}     & \textbf{73.8 (+3.4)}    & \textbf{62.3 (+5.2)}   & \textbf{69.1 (+4.1)} 	\\
\bottomrule
\end{tabular}%
}
\vspace{-.1in}
\caption{Results on the 10-10 setting of Pascal VOC. $^{\ast}$ denotes the results are from~\cite{maracani2021recall}, $^{\dagger}$ denotes the results are from~\cite{cermelli2022wilson}.}
\label{tbl:voc10-10}
\end{table}

\subsubsection{Performance on COCO-to-VOC}
This is a more challenging setting. First, the base model is trained on 60 classes of the COCO dataset, which are not overlapped with VOC. Then we train the additional 20 classes from VOC dataset with image-level labels in the second step.
The results are reported in Table~\ref{tbl:coco2voc}, \Ours{} also achieves the new
state-of-the-art performance against the WILSS methods.

\begin{table}[t!]
\centering
\resizebox{0.85\linewidth}{!}{%
\begin{tabular}{rc|lll|l}
\toprule
\multirow{2}{*}{Method} & \multirow{2}{*}{Sup}      & \multicolumn{3}{c|}{COCO} &  VOC \\
                        &                           & 1-60    & 61-80    & All     & 61-80        \\ 
\midrule
FT$^{\dagger}$                         			& \pixel{}	& 1.9     	& 41.7     & 12.7	& \underline{75.0}     \\
LWF$^{\dagger}$~\cite{li2017lwf}       		& \pixel{}      & 36.7	& \underline{49.0}     & \underline{40.3}    	& 73.6       \\
ILT$^{\dagger}$~\cite{michieli2019ilt}      		& \pixel{}    	& \underline{37.0}    	& 43.9     & 39.3    	& 68.7      \\
MIB$^{\dagger}$~\cite{cermelli2020mib}    	& \pixel{}    	& 34.9    	& 47.8     & 38.7    	& 73.2       \\
PLOP$^{\dagger}$~\cite{douillard2021plop}            	& \pixel{}    	& 35.1    	& 39.4     & 36.8    	& 64.7      \\
\midrule
CAM$^{\dagger}$                     					& \image{}   & 30.7    	& 20.3     & 28.1    	& 39.1     \\
SEAM$^{\dagger}$~\cite{wang2020seam}        		& \image{}   & 31.2    	& 28.2     & 30.5    	& 48.0       \\
SS$^{\dagger}$~\cite{araslanov2020ss}       			& \image{}   & 35.1    	& 36.9     & 35.5    	& 52.4      \\
EPS$^{\dagger}$~\cite{lee2021eps}           			& \image{}   & 34.9    	& 38.4     & 35.8    	& 55.3      \\
WILSON$^{\dagger}$~\cite{cermelli2022wilson}  		& \image{}   & 39.8    	& 41.0     & 40.6    	& 55.7      \\ %\hline
\Ours{} (Ours)                  		& \image{}   & \textbf{39.9 (+0.1)}   & \textbf{44.7 (+3.7)} & \textbf{41.6 (+1.0)}    & \textbf{63.6 (+7.9)}      \\
\bottomrule
\end{tabular}%
}
\vspace{-.1in}
\caption{Results on the COCO-to-VOC setting. $^{\dagger}$ denotes the results are from~\cite{cermelli2022wilson}.}
\label{tbl:coco2voc}
\vspace{-.1in}
\end{table}

\subsection{Ablation Studies}
%In this section, we present ablation studies to empirically justify the design choices. 
Unless otherwise specified, we perform the ablation experiments with iBOT~\cite{zhou2021ibot} as the self-supervised model with ViT-B architecture based on the 10-10 VOC setting.
We provide more ablation details, \eg, ablations of $\alpha, \beta, \mathcal{K}$, in our ${\tt supplementary}$ materials.

\subsubsection{Effect of Number of Seeds}
\begin{figure}[t!]
\centering
	\includegraphics[width=0.94\linewidth]{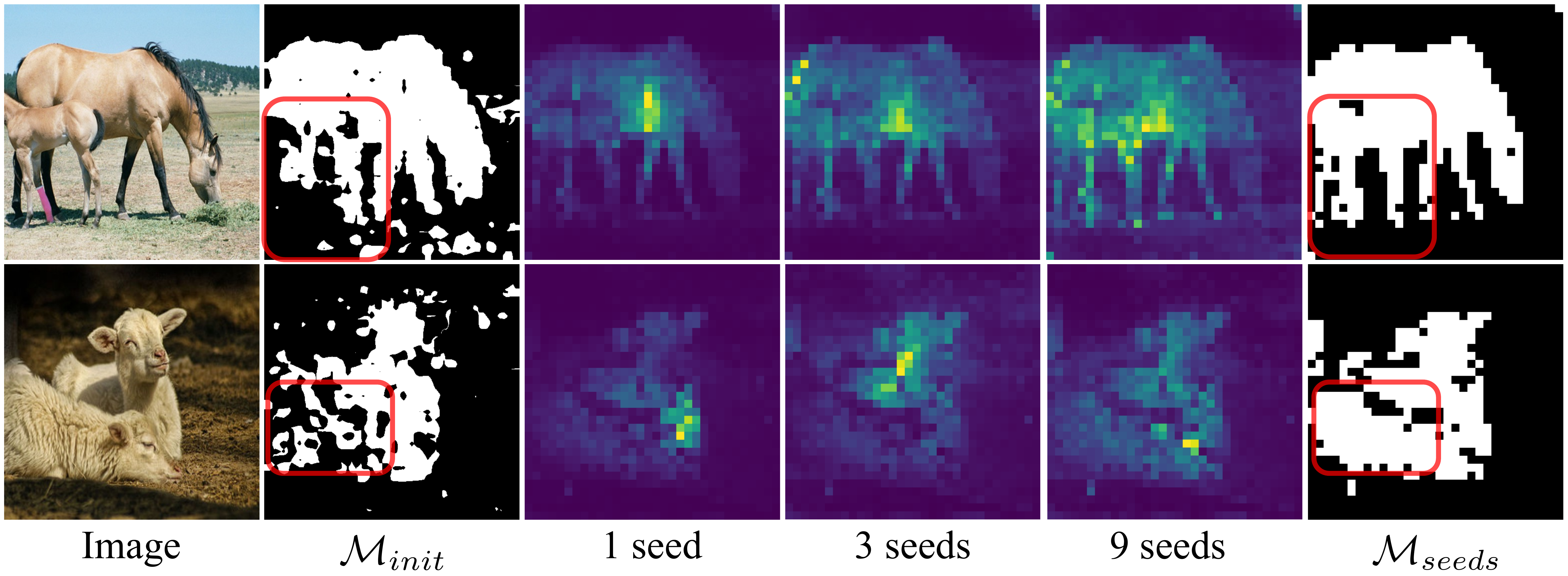}
    \vspace{-.1in}
	\caption{Effect of the number of randomly selected seeds.}
	\label{fig:seeds}
	\vspace{-.1in}
\end{figure}

%可以给seed个数对attention map可视化
As depicted in Figure~\ref{fig:seeds}, we visualize the impact of number of seeds when calculating $\mathcal{M}_{seeds}$. Based on the category-aware $\mathcal{M}_{init}$ of two classes (\textit{horse, dog}), more random seeds guidance leads to more compact attention maps and nine seeds are enough to get good clustering results for training efficiency.
Moreover, the category-agnostic $\mathcal{M}_{seeds}$ can be complementary to the initial $\mathcal{M}_{init}$, which is indicated by red boxes.

\subsubsection{Analysis of Mask Fusion}
%单独maskclip
%maskclip 并 ibot
%maskclip 交 ibot
In this section, we verify the effect of mask fusion operation in Eq.~(\ref{eq:fusion}). As shown in Table~\ref{tbl:mask_fusion}, we compare the results of two fusion operations and the result without fusion (``None''). Performing co-segmentation with ``Union'' can significantly improve performance, especially on new classes (53.25\% $\rightarrow$ 55.45\%, 56.39\% $\rightarrow$ 60.54\%).
%We further analyze the impact of the $\mathcal{K}$ value when binarize $\mathcal{M}_{seeds}$, as shown in Table~\ref{tbl:mask_unionpercent}, preserving too little ($\mathcal{K}=90\%$) information from the self-supervised model can not utilize the complementary knowledge effectively, while keeping too much ($\mathcal{K}=60\%$) will introduce more noise. We thus set $\mathcal{K}=70\%$ for all other experiments.
The results indicate that the proposed pre-training based co-segmentation does improve the supervision and performance for WILSS against original image-level labels.

\begin{table}[t!]
\centering
\resizebox{0.8\linewidth}{!}{%
\begin{tabular}{c|ccc|ccc}
\toprule
 Fusion 		& \multicolumn{3}{c|}{Disjoint} & \multicolumn{3}{c}{Overlap} \\
Operation    & 1-10    & 11-20    & All     & 1-10    & 11-20    & All    \\ 
\midrule
None           &  67.83           &  53.25        &  61.96       &  73.25      & 56.39        & 66.45  \\
Intersection	       &  66.85           & 51.17         &  60.40       &  74.12      & 54.02        & 64.95  \\
Union                      &  67.23           &  55.45        &  \textbf{62.74}       &  73.12      & 60.54        & \textbf{67.94}  \\

\bottomrule
\end{tabular}%
}
\vspace{-.1in}
\caption{Performance comparison of the mask fusion operation. ``None'' denotes using only the $\mathcal{M}_{init}$ from CLIP.}
\label{tbl:mask_fusion}
\end{table}

% \begin{table}[t!]
% \centering
% \resizebox{0.8\linewidth}{!}{%
% \begin{tabular}{c|ccc|ccc}
% \toprule
% \multirow{2}{*}{ $\mathcal{K}$ (\%)} 		& \multicolumn{3}{c|}{Disjoint} & \multicolumn{3}{c}{Overlap} \\
%          						        & 1-10    & 11-20    & All     & 1-10    & 11-20    & All    \\ 
% \midrule

% 90           & 67.74         &    54.38      &  62.46     &   74.14      &   59.00      & 67.67  \\
% 80	       & 67.52         &   55.08      &  62.70     &   73.73      &   59.94      & 67.95  \\
% 70           & 67.23         &   55.45      &  \textbf{62.74}      & 73.12        & 60.54        & \textbf{67.95}  \\
% 60           & 66.99	   &   55.48      &  62.63	&  72.51	   &  60.49	       &  67.63 \\
% %50           &             &          &         &         &          &  \\
% \bottomrule
% \end{tabular}%
% }
% \caption{Ablation experiments on the mask union operation by varying the percentage value $\mathcal{K}$.}
% \label{tbl:mask_unionpercent}
% \end{table}

\subsubsection{Effect of Dense Contrastive Objective}
%加不加，以及loss weight影响
We further analyze the effect of the proposed dense contrastive loss in Table~\ref{tbl:contrast}. Compared to the result without using $\mathcal{L}^{new}_{\text{DCL}}$ ($\lambda=0.0$), introducing our proposed dense contrastive loss with a loss weight of 0.1 further improves the performance on new classes by 1.77\% and 0.38\% in the two settings, respectively. We thus set $\lambda=0.1$ in our experiments by default.
The results show that the plug-in teacher module equipped with the proposed $\mathcal{L}^{new}_{\text{DCL}}$ can more effectively optimize and leverage the dense supervision for WILSS.

\begin{table}[t!]
\centering
\resizebox{0.8\linewidth}{!}{%
\begin{tabular}{c|ccc|ccc}
\toprule
\multirow{2}{*}{$\lambda$ of $\mathcal{L}^{new}_{\text{DCL}}$} 		& \multicolumn{3}{c|}{Disjoint} & \multicolumn{3}{c}{Overlap} \\
                                                                & 1-10    & 11-20    & All     & 1-10    & 11-20    & All    \\ 
\midrule
0.0     & 67.23         &   55.45      &  62.74         & 73.12        & 60.54        & 67.95    \\
0.1     & 67.94         &   57.22      &  \textbf{63.91}        &  73.44        &  60.92       & \textbf{68.28}   \\
1.0     & 67.15         &   57.94      &  63.87         & 72.08        &  61.05       &  68.17 \\							     
\bottomrule
\end{tabular}%
}
\vspace{-.1in}
\caption{Ablation experiments on the dense contrastive objective $\mathcal{L}^{new}_{\text{DCL}}$. $\lambda=0.0$ represents eliminating the dense contrastive loss.}
\label{tbl:contrast}
%\vspace{-.1in}
\end{table}

\subsubsection{Effect of Memory-based Copy-paste Augmentation}
%加不加，以及memory-bank的影响, 以及加的随机概率影响
As reported in Table~\ref{tbl:copypaste}, we analyze the effect of the proposed memory-based copy-paste augmentation. Without further tuning, we fix the augmentation probability as 0.5.
Compared with the results without augmentation, applying the copy-paste augmentation with $\mathcal{B}=50$ can already improve the performance of old classes (67.94\% $\rightarrow$ 68.64\% in disjoint, 73.44\% $\rightarrow$ 73.71\% in overlap) for saving memory consumption.

\begin{table}[t!]
\centering
\resizebox{0.8\linewidth}{!}{%
\begin{tabular}{cc|ccc|ccc}
\toprule
\multirow{2}{*}{Rand.}		& Memory-bank 		& \multicolumn{3}{c|}{Disjoint} & \multicolumn{3}{c}{Overlap} \\
                            & $\mathcal{B}$   	& 1-10    & 11-20    & All     & 1-10    & 11-20    & All    \\ 
\midrule
\multirow{5}{*}{0.5}  	  &  0      & 67.94         	&   57.22      	&  63.91       &   73.44       &  60.92       & 68.28   \\
					   &  10       & 68.85            	&   57.14       	&  64.29       &    73.11      &  61.50       & 68.40  \\
						 &  50       & 68.64           	& 57.90         	&  \textbf{64.56}       &    73.71     	&  62.06       & \textbf{68.95}	  \\
						 & 100     & 68.41   		& 58.09       	&  64.55       &    73.26     	&  61.78       & 68.60   \\
						 &  500     & 68.73           	& 58.29        	&  64.80       &    73.50      &  62.28       & 68.97   \\ 
						   % \cmidrule{2-9}
						   % & 0.1  				&  \multirow{3}{*}{50?}      &             &          &         &         &          &  \\
						   % & 0.5  				&       				&             &          &         &         &          &  \\					
						   % & 1.0  				&       				&             &          &         &         &          &  \\		     
\bottomrule
\end{tabular}%
}
\vspace{-.1in}
\caption{Ablation experiments on the memory-based copy-paste augmentation. $\mathcal{B}=0$ denotes removing augmentation.}
\label{tbl:copypaste}
\vspace{-.1in}
\end{table}

\subsubsection{Effect of Different Self-supervised Models}
%比较不同的自监督模型， ibot系列，和DINO
We experiment with two representative self-supervised models~\cite{zhou2021ibot,caron2021dino} based on different ViT architectures.
As shown in Table~\ref{tbl:selfup_model}, both bring promising results and we use iBOT with ViT-L for better and stable performance.% in our framework.

\begin{table}[t!]
\centering
\resizebox{0.85\linewidth}{!}{%
\begin{tabular}{cc|ccc|ccc}
\toprule
Self-supervised		& \multirow{2}{*}{Arch.}  & \multicolumn{3}{c|}{Disjoint} & \multicolumn{3}{c}{Overlap} \\
Model				& 			     		  & 1-10    & 11-20    & All     & 1-10    & 11-20    & All    \\ 
\midrule
\multirow{2}{*}{iBOT~\cite{zhou2021ibot}}  	  &  ViT-B  & 68.64  & 57.90  &  64.56  & 73.71 &  62.06 & 68.95	  \\
						                   &  ViT-L  & 68.51  & 58.20  &  \textbf{64.64} &  73.79 & 62.26  & \textbf{69.09} \\
						                     \cmidrule{2-8}
\multirow{2}{*}{DINO~\cite{caron2021dino}} 	  &  ViT-B  & 68.01  & 59.14  &  64.86  & 73.11 &  60.98 & 68.14  \\
						                     &  ViT-S  & 66.86  & 58.85	&  64.17  & 70.97 &	 59.57 & 66.43  \\												     
\bottomrule
\end{tabular}%
}
\vspace{-.1in}
\caption{Performance comparison by using different self-supervised pre-training models.}
\label{tbl:selfup_model}
\end{table}

\subsubsection{Factor-by-factor Experiments}
We conduct a factor-by-factor experiment on the proposed pre-training based co-segmentation, dense contrastive loss, and memory-based copy-paste augmentation.
As shown in Tabel~\ref{tbl:factorbyfactor}, each design has a positive effect, and all designs are combined to obtain the best performance.

\begin{table}[t!]
\centering
\resizebox{\linewidth}{!}{%
\begin{tabular}{ccc|ccc|ccc}
\toprule
Pre-training         & \multirow{2}{*}{$\mathcal{L}^{new}_{\text{DCL}}$}  & Copy-paste	    & \multicolumn{3}{c|}{Disjoint} & \multicolumn{3}{c}{Overlap} \\
based Co-seg.        &                                             & Aug.           & 1-10  & 11-20  & All   & 1-10  & 11-20 & All  \\ 
\midrule
                     &                                             &                & 67.83 & 53.25  & 61.96  & 73.25 & 56.39 & 66.45 \\
\checkmark           &                                             &                & 67.23 & 55.45  & 62.74  & 73.12 & 60.54 & 67.94 \\
\checkmark           &  \checkmark                                 &                & 67.94 & 57.22  & 63.91  & 73.44 & 60.92 & 68.28 \\
\checkmark           &  \checkmark                                 & \checkmark     & 68.64 & 57.90  & 64.56  & 73.71 & 62.06 & 68.95 \\
\checkmark           &  \checkmark                                 & \checkmark     & 68.51 & 58.20  & \textbf{64.64}  & 73.79 & 62.26 & \textbf{69.09} \\
\bottomrule
\end{tabular}%
}
\vspace{-.1in}
\caption{Effect of each design proposed in this work. The last row indicates the result of using ViT-L architecture.}
%\vspace{-.1in}
\label{tbl:factorbyfactor}
\end{table}

\begin{table}[t!]
\centering
\resizebox{0.8\linewidth}{!}{%
\begin{tabular}{rc|ccc|ccc}
\toprule
\multirow{2}{*}{Method}		        & \multirow{2}{*}{Sup} 		 & \multicolumn{3}{c|}{Disjoint} & \multicolumn{3}{c}{Overlap} \\
                                    &    	                     & 1-10    & 11-20    & All     & 1-10    & 11-20    & All    \\ 
\midrule
WILSON~\cite{cermelli2022wilson}    & I                          & 64.5 & 54.3 & 60.8 & 70.4 & 57.1 & 65.0 \\  
WILSON~\cite{cermelli2022wilson}    & P                          & 69.5 & 56.4 & 64.2 & 73.6 & 57.6 & 66.7 \\
\midrule
\Ours{} (Ours)  	                &  I                         & 68.5 & 58.2 & \textbf{64.6} & 73.8 & 62.3 & \textbf{69.1}  \\
\bottomrule
\end{tabular}%
}
\vspace{-.1in}
\caption{Comparison with previous WILSS SoTA~\cite{cermelli2022wilson} trained with direct dense annotations.}
\label{tbl:com_dense}
\end{table}

\begin{table}[t!]
\centering
\resizebox{0.8\linewidth}{!}{%
\begin{tabular}{rc|ccc|ccc}
\toprule
\multirow{2}{*}{Method}             & Train  		 & \multicolumn{3}{c|}{Disjoint} & \multicolumn{3}{c}{Overlap} \\
                                    & Data    	     & 1-10    & 11-20    & All     & 1-10    & 11-20    & All    \\ 
\midrule
WILSON~\cite{cermelli2022wilson}    & 100\%          & 64.5    & 54.3  & 60.8  & 70.4  & 57.1  & 65.0   \\
\midrule
\multirow{3}{*}{\Ours{} (Ours)}            & 100\%          & 68.5    & 58.2  & 64.6  & 73.8  & 62.3  & 69.1	\\ 
                                    %& 75\%           & 68.8	   & 57.5  & 64.4  &       &       &  \\  
                                    & 50\%           & 66.7    & 56.0  & 62.7  & 72.1  & 60.5  & 67.4 \\  
                                    & 30\%           & 68.5    & 51.5  & 61.5  & 75.7  & 55.7  & 66.8 \\  
\bottomrule
\end{tabular}%
}
\vspace{-.1in}
\caption{Performance comparison with fewer training data.}
\vspace{-.1in}
\label{tbl:datasize}
\end{table}

\subsubsection{Data Efficiency Experiment}
%数据高效实验, 结合supp里的收敛性分析
We first compare with WILSON~\cite{cermelli2022wilson} trained with direct dense pixel-level labels (P) for all classes in Table~\ref{tbl:com_dense}. It is worth noting that \Ours{} based on only image-level labels (I) still outperforms it, especially in the challenging overlap setting.
We further evaluate \Ours{} with fewer training data in Table~\ref{tbl:datasize}. It is notable that \Ours{} trained with only 30\% data still outperforms WILSON trained with 100\% data.
Combining Table~\ref{tbl:com_dense}, Table~\ref{tbl:datasize}, and the fast convergence result (Sec.~4) in {\tt supplementary} materials, we can conclude that \Ours{} is a significantly data-efficient WILSS framework.

\subsubsection{Visualization}

\begin{figure}[t!]
\centering
	\includegraphics[width=0.91\linewidth]{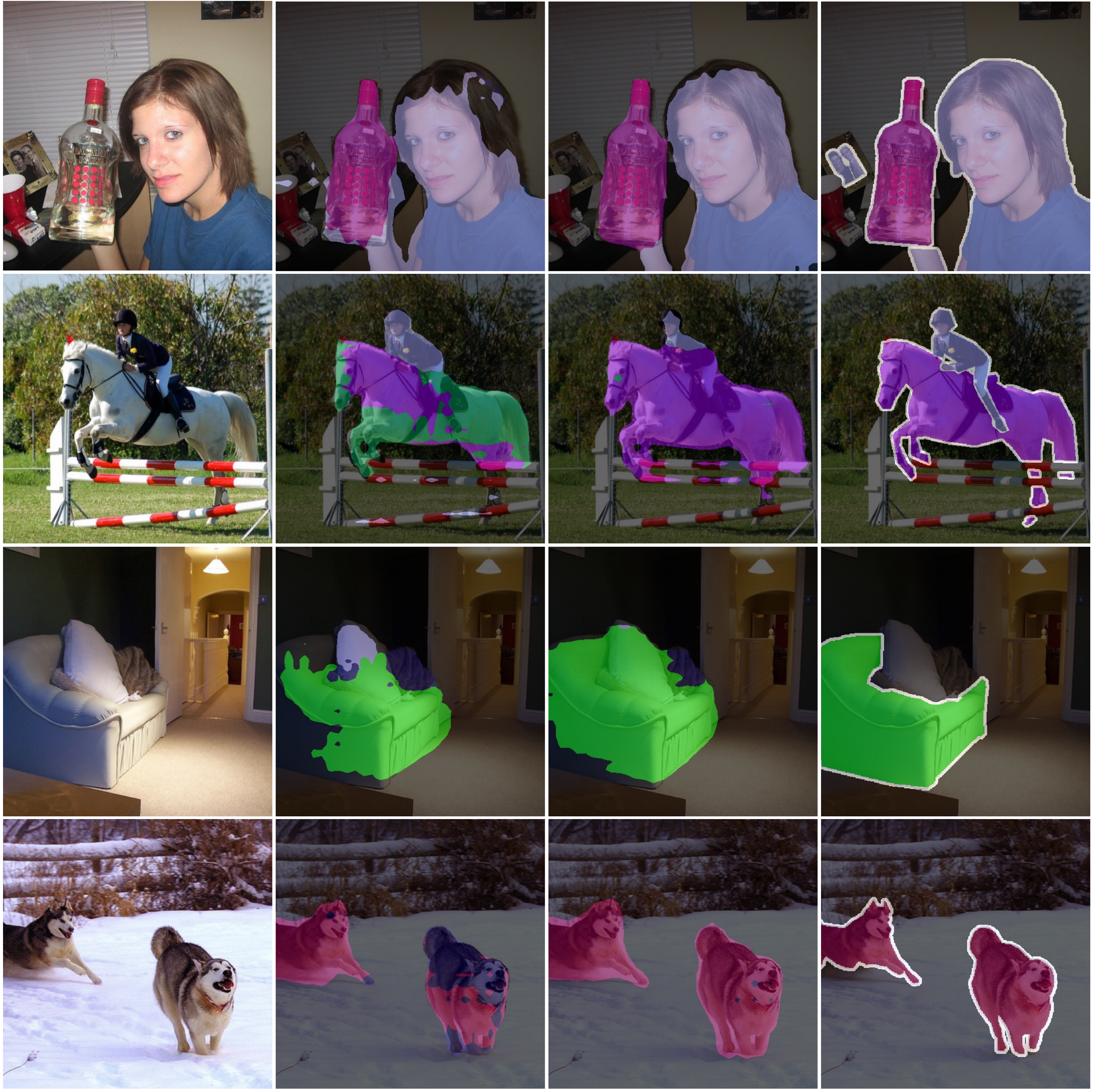}
    \vspace{-.1in}
	\caption{Qualitative comparison between previous WILSS SoTA~\cite{cermelli2022wilson} and our \Ours{} on the 10-10 VOC setting. From left to right: image, WILSON~\cite{cermelli2022wilson}, \Ours{} and the ground-truth.}
	\label{fig:visualize}
	\vspace{-.1in}
\end{figure}

We report qualitative results indicating the superiority of our \Ours{} framework on both old (\eg, \textit{bottle}) and new (\eg, \textit{person, horse, dog, sofa}) classes in Figure~\ref{fig:visualize}.

%--------------------Liminations----------------------------------------------------------
\section{Limitations}
Like WILSON~\cite{cermelli2022wilson}, the \Ours{} framework is not designed for single-class incremental learning steps, the proposed dense contrastive objective needs some other classes as negative samples.

%--------------------Conclusion----------------------------------------------------------
\section{Conclusion}

In this paper, we present a novel and data-efficient \Ours{} framework for weakly incremental learning for semantic segmentation. \Ours{} is the first attempt to exploit knowledge of pre-trained foundation models for WILSS. 
We propose pre-training based co-segmentation to generate dense supervision based on image-level labels.
We further use a teacher-student architecture with a proposed dense contrastive loss to more effectively utilize the pseudo labels.
Besides, we introduce memory-based copy-paste augmentation to improve the forgetting problem of old classes.
Extensive experiments demonstrate the superior results of \Ours{}.

%%%%%%%%% REFERENCES
{\small
\bibliographystyle{ieee_fullname}
\bibliography{egbib}
}

\newpage

\ifx\AddAppendix\undefined
\else
\appendix
\section{Appendix}

\subsection{Effect of $\alpha$ and $\beta$}
%union 70, CL 0.1, copypaste 0.5-50, alpha=0.5默认
We analyze the sensitivity of $\alpha$ and $\beta$ and report the results in Table~\ref{tbl:alpha_beta}.
First, we fix $\beta=1.0$ and find that equally ($\alpha=0.5$) integrating the hard and soft dense labels for new classes from the teacher can better improve the performance from 62.15\% to 64.21\%.
Then, we fix $\alpha=0.5$ and find that paying more attention ($\beta=0.9$) to the outputs of old model $f^{t-1}_{\theta}$ than the teacher when supervise the old classes can achieve better overall performance.
We set $\alpha=0.5, \beta=0.9$ in all our experiments by default.

\begin{table}[h!]
\centering
\resizebox{0.8\linewidth}{!}{%
\begin{tabular}{lcc|ccc}
\toprule
Method 		 			  & $\alpha$		& $\beta$		&  1-10    & 11-20    & All   \\
\midrule
\multirow{8}{*}{\Ours{}}      	&  0.0				&   \multirow{3}{*}{1.0}      	&  67.19           &  54.30        &  62.15       \\
						&  0.5				&         				&  68.11           &  57.67        &  64.21        \\
						&  1.0				&         				&  67.20           &  57.64        &  64.13     \\
						\cmidrule{2-6}
						& \multirow{4}{*}{0.5} 	&  0.5     				&  59.17           &  56.87        &  59.55        \\					
						&   					&  0.7     				&  64.73           &  57.23        &  62.37         \\
						%&   					&  0.9     				&  67.33          &   59.04       &   \textbf{64.49}        \\
                        &   					&  0.9     				&  68.64          &   57.90       &   \textbf{64.56}        \\
                        &   					&  1.0     				&  68.11           &  57.67        &  64.21        \\
							     
\bottomrule
\end{tabular}%
}
\caption{Ablation experiments on the parameters $\alpha$ and $\beta$ based on the disjoint protocol.}
\label{tbl:alpha_beta}
\end{table}

\subsection{Number of Points in Dense Contrastive Loss}
We analyze our \Ours{} with different number of randomly sampled points when calculating the proposed $\mathcal{L}^{new}_{\text{DCL}}$.
The results are reported in Table~\ref{tbl:points_dcl}.
We can see that \Ours{} is robust to the number of sample points, and randomly sampling more points (\eg, 50) for each class may introduce more noise and thus degrade the performance. We randomly sample ten points when calculating $\mathcal{L}^{new}_{\text{DCL}}$ in our experiments by default.

\begin{table}[t!]
\centering
\resizebox{1.0\linewidth}{!}{%
\begin{tabular}{c|ccc|ccc}
\toprule
Points in 		 & \multicolumn{3}{c|}{Disjoint} & \multicolumn{3}{c}{Overlap} \\
$\mathcal{L}^{new}_{\text{DCL}}$   	         & 1-10    & 11-20    & All     & 1-10    & 11-20    & All    \\ 
\midrule
%0        & 67.23  & 55.45  & 62.74  & 73.12  & 60.54  & 67.95 \\  for iBOT-ViT-Base
5        & 68.37  & 58.18  & 64.56  & 73.64  & 62.38  & 69.04  \\
10       & 68.51  & 58.20  & \textbf{64.64}  & 73.79  & 62.26  & \textbf{69.09}  \\
50       & 68.67  & 57.72  & 64.49  & 73.35  & 61.67  & 68.58  \\
\bottomrule
\end{tabular}%
}
\caption{Ablation experiments on the number of points sampled for each contained class of the proposed dense contrastive loss.}
\label{tbl:points_dcl}
\end{table}

\subsection{Analysis of Mask Fusion}
We analyze the impact of the $\mathcal{K}$ value when binarize $\mathcal{M}_{seeds}$, as shown in Table~\ref{tbl:mask_unionpercent}, preserving too little ($\mathcal{K}=90\%$) information from the self-supervised model can not utilize the complementary knowledge effectively, while keeping too much ($\mathcal{K}=60\%$) will introduce more noise. We thus set $\mathcal{K}=70\%$ for all other experiments.

\begin{table}[t!]
\centering
\resizebox{0.8\linewidth}{!}{%
\begin{tabular}{c|ccc|ccc}
\toprule
\multirow{2}{*}{ $\mathcal{K}$ (\%)} 		& \multicolumn{3}{c|}{Disjoint} & \multicolumn{3}{c}{Overlap} \\
         						        & 1-10    & 11-20    & All     & 1-10    & 11-20    & All    \\ 
\midrule

90           & 67.74         &    54.38      &  62.46     &   74.14      &   59.00      & 67.67  \\
80	       & 67.52         &   55.08      &  62.70     &   73.73      &   59.94      & 67.95  \\
70           & 67.23         &   55.45      &  \textbf{62.74}      & 73.12        & 60.54        & \textbf{67.95}  \\
60           & 66.99	   &   55.48      &  62.63	&  72.51	   &  60.49	       &  67.63 \\
%50           &             &          &         &         &          &  \\
\bottomrule
\end{tabular}%
}
\caption{Ablation experiments on the mask union operation by varying the percentage value $\mathcal{K}$.}
\label{tbl:mask_unionpercent}
\end{table}

\begin{figure}[t!]
\centering
	\includegraphics[width=0.9\linewidth]{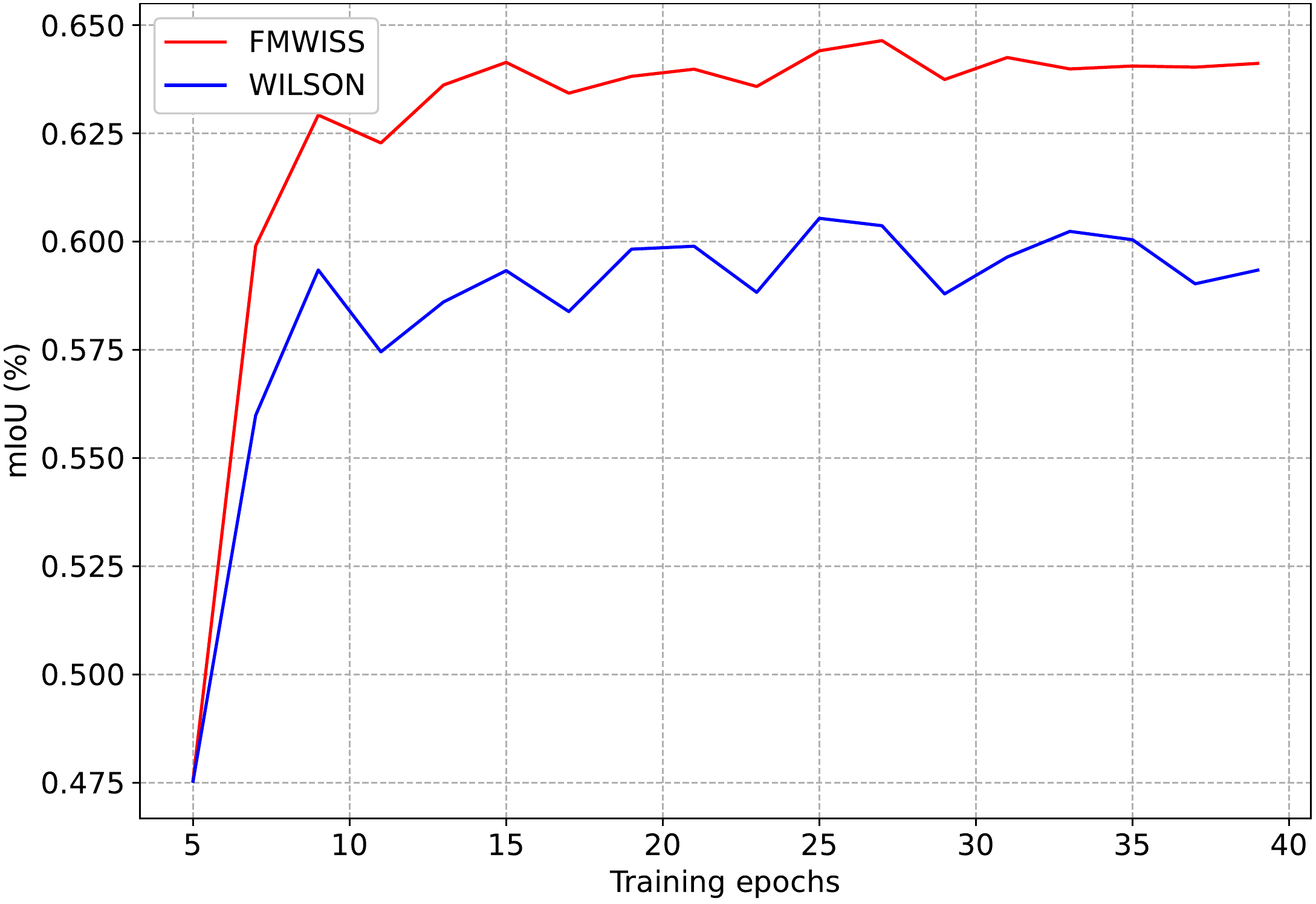}
    \vspace{-.1in}
	\caption{Convergence comparison between previous WILSS SoTA~\cite{cermelli2022wilson} and our \Ours{} based on 10-10 VOC setting.}
	\label{fig:convergence}
\end{figure}

\begin{figure*}[t!]
\centering
	\includegraphics[width=1.0\linewidth]{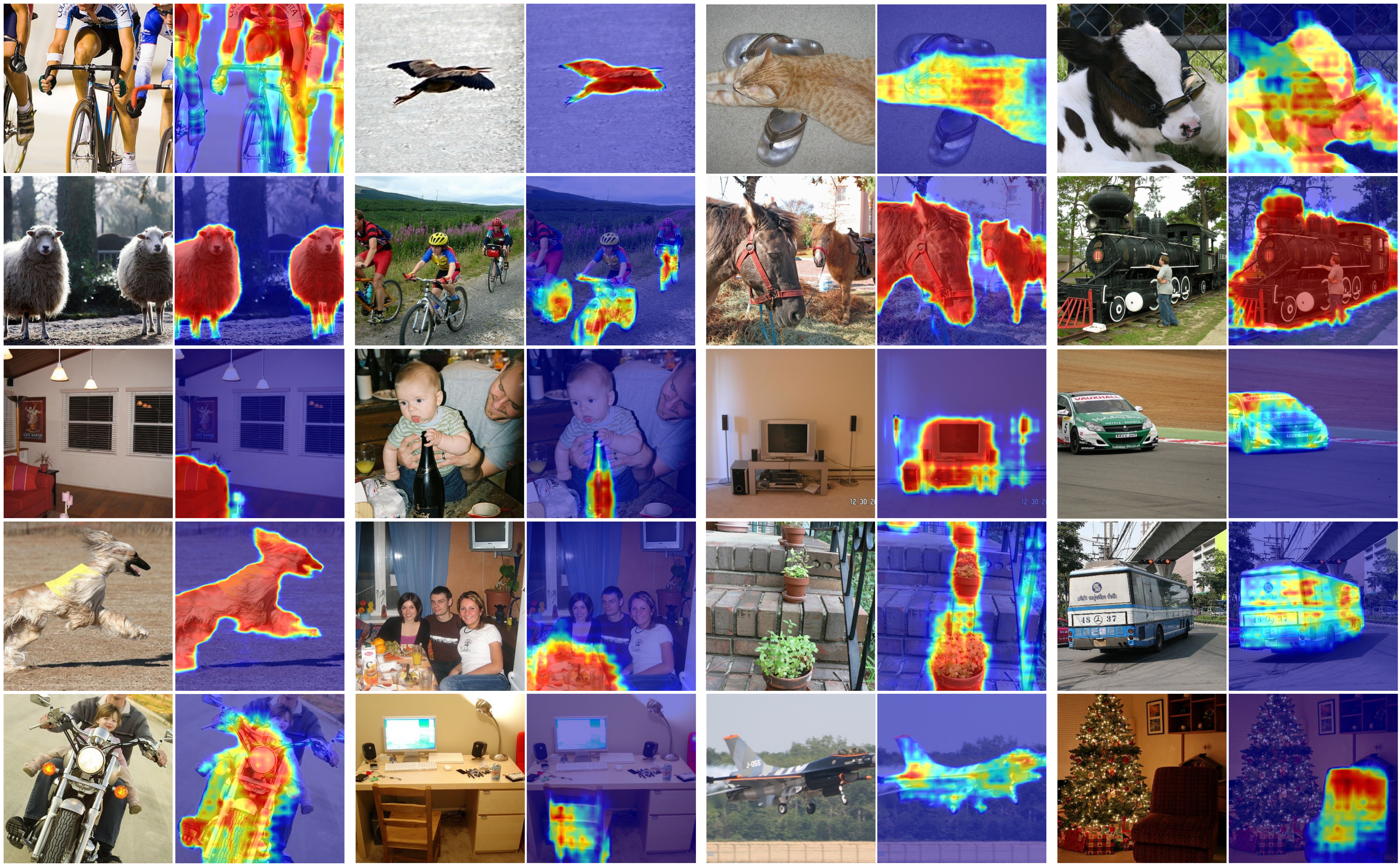}
    \vspace{-.1in}
	\caption{Qualitative visualization of the learned features maps of the teacher module on 10-10 VOC setting.}
	\label{fig:supp_vis_teacher}
\end{figure*}

\begin{figure*}[t!]
\centering
	\includegraphics[width=1.0\linewidth]{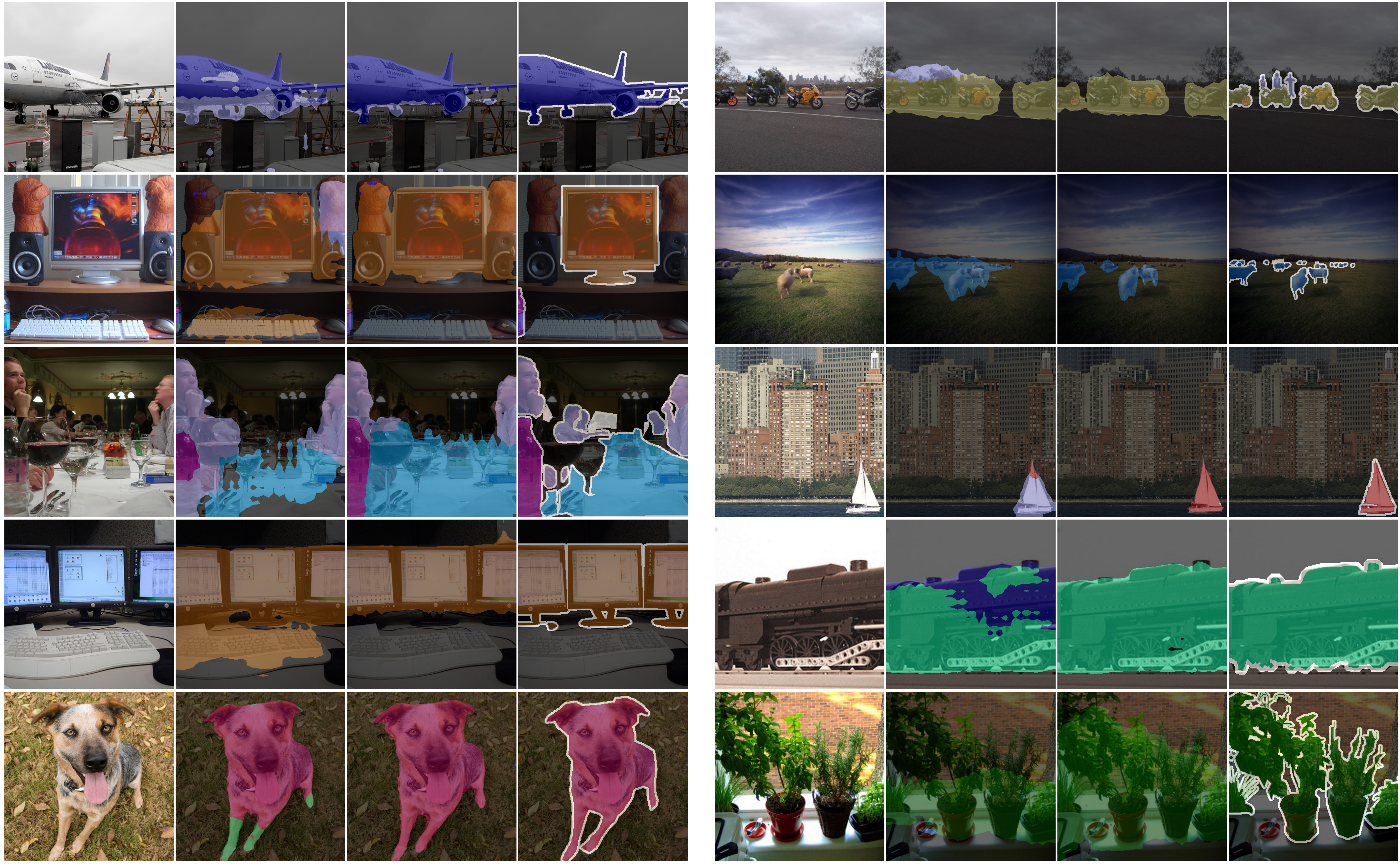}
    \vspace{-.1in}
	\caption{More qualitative comparison between previous WILSS SoTA~\cite{cermelli2022wilson} and our \Ours{} on the 10-10 VOC setting. From left to right: image, WILSON~\cite{cermelli2022wilson}, \Ours{} (ours) and the ground-truth.}
	\label{fig:supp_vis_seg}
\end{figure*}

\subsection{Analysis of Convergence}  %正文提了Sec.4， 不要换
%探究我们的方法和WILSON的收敛性分析比较
As shown in Figure~\ref{fig:convergence}, we report the convergence curve \wrt~training process on the 10-10 VOC setting. We can conclude that our \Ours{} can converge faster and can consistently improve the performance compared to the previous SoTA~\cite{cermelli2022wilson}.

%可以放到supp里的：
%\subsubsection{Effectiveness of Sample Number of DCL}
%探究DCL中采样点个数的影响，最后做

% \section{Comparison with Using Dense Supervision for All Classes}
% %对比wilson里给出的全用pixel gt的结果，我们的仅用image-level的还是高的
% As shown in Table~\ref{tbl:com_dense}, we compare the performance of other WILSS methods when the dense labels are provided for both old and new classes.
% It is worth noting that our \Ours{} based on image-level labels can still outperform others, especially in the challenging overlap setting.

% \begin{table}[t!]
% \centering
% \resizebox{1.0\linewidth}{!}{%
% \begin{tabular}{cc|ccc|ccc}
% \toprule
% \multirow{2}{*}{Method}		        & \multirow{2}{*}{Sup} 		 & \multicolumn{3}{c|}{Disjoint} & \multicolumn{3}{c}{Overlap} \\
%                                     &    	                     & 1-10    & 11-20    & All     & 1-10    & 11-20    & All    \\ 
% \midrule
% WILSON~\cite{cermelli2022wilson}    & I                          & 64.5 & 54.3 & 60.8 & 70.4 & 57.1 & 65.0 \\  
% WILSON~\cite{cermelli2022wilson}    & P                          & 69.5 & 56.4 & 64.2 & 73.6 & 57.6 & 66.7 \\
% \midrule
% \Ours{} (Orus)  	                &  I                         & 68.5 & 58.2 & 64.6 & 73.8 & 62.3 & 69.1  \\
% \bottomrule
% \end{tabular}%
% }
% \caption{Comparison with other weakly supervised methods trained with direct dense labels.}
% \label{tbl:com_dense}
% \end{table}

\subsection{Visualization}
To analyze the effect of the plug-in teacher module, we visualize the learned feature maps for all the old classes (\textit{aeroplane, bicycle, bird, boat, bottle, bus, car, cat, chair, cow}) and all the new classes (\textit{dining table, dog, horse, motorbike, person, potted plant, sheep, sofa, train, tv monitor}) based on the 10-10 VOC setting. As shown in Figure~\ref{fig:supp_vis_teacher}, it can be seen that the teacher module of \Ours{} framework can learn clear and high attention scores for different correct classes.

As depicted in Figure~\ref{fig:supp_vis_seg}, we show more qualitative results indicating the superiority of our \Ours{} framework on both old (\eg, \textit{aeroplane, bottle, boat}) and new (\eg, \textit{dog, motorbike, person, dining table, tv monitor, sheep, train, potted plant}) classes.

\end{document}